\documentclass{sig-alternate}
\newfont{\mycrnotice}{ptmr8t at 7pt}
\newfont{\myconfname}{ptmri8t at 7pt}
\let\crnotice\mycrnotice%
\let\confname\myconfname%

\usepackage[german,american]{babel}
\usepackage[T1]{fontenc}
\usepackage[utf8]{inputenc}
\usepackage{times}
\usepackage{textcomp}
\usepackage{graphicx}
\usepackage{amssymb}
\usepackage{microtype} 

\usepackage{url}
\renewcommand{\tt}{\fontencoding{OT1}\fontfamily{cmtt}\selectfont}

\usepackage[usenames,dvipsnames]{color}
\usepackage{booktabs}
\usepackage[numbers,sectionbib]{natbib}
\makeatletter
\def\@copyrightspace{\relax}
\makeatother

\begin{document}

\title{Fishing for Clickbaits in Social Images and Texts with Linguistically-Infused Neural Network Models}
\subtitle{The Pineapplefish Clickbait Detector at the Clickbait Challenge 2017}

\numberofauthors{2}
\author{
\alignauthor
Maria Glenski\\
\affaddr{University of Notre Dame\\ Notre Dame, IN 46556 USA}\\
\affaddr{mglenski@nd.edu}\\
\alignauthor
Ellyn Ayton\\
\affaddr{Western Washington University\\ Bellingham, WA 98225 USA}\\
\affaddr{aytone@wwu.edu}\\
\and 
\alignauthor
Dustin Arendt\\
\affaddr{Visual Analytics Group\\ National Security Directorate \\
  Pacific Northwest National Laboratory\\
  902 Battelle Blvd, Richland, WA 99354 }\\  
\affaddr{dustin.arendt@pnnl.gov}\\
\alignauthor
Svitlana Volkova\\
\affaddr{Data Sciences and Analytics Group\\ National Security Directorate \\
  Pacific Northwest National Laboratory\\
  902 Battelle Blvd, Richland, WA 99354 }\\ 
\affaddr{svitlana.volkova@pnnl.gov}\\
}

\maketitle

\begin{abstract} 
This paper presents the results and conclusions
of our participation in the Clickbait Challenge 2017\footnote{http://www.clickbait-challenge.org/} on automatic clickbait detection in social media. We first describe linguistically-infused neural network models and identify informative representations to predict the level of clickbaiting present in Twitter posts. Our models allow to answer the question not only whether a post is a clickbait or not, but to what extent it is a clickbait post \textit{e.g.}, not at all, slightly, considerably, or heavily clickbaity using a score ranging from 0 to 1. We evaluate the predictive power of models trained on varied text and image representations extracted from tweets. Our best performing model that relies on the tweet text and linguistic markers of biased language extracted from the tweet and the corresponding page yields mean squared error (MSE) of 0.04, mean absolute error (MAE) of 0.16 and R2 of 0.43 on the held-out test data. For the binary classification setup (clickbait vs. non-clickbait), our model achieved F1 score of 0.69. We have not found that image representations combined with text yield significant performance improvement yet. Nevertheless, this work is the first to present preliminary analysis of  objects extracted using Google Tensorflow object detection API from images in clickbait vs. non-clickbait Twitter posts. Finally, we outline several steps to improve model performance as a part of the future work.
 
\end{abstract}

\section{Introduction}
Clickbait posts (aka eye-catching headlines) are designed to lure readers into clicking associated links by exploiting curiosity with vague, exaggerated, sensational and misleading content. The intent behind clickbait messages varies from attention redirection to monetization and traffic attraction. Automatic detection of clickbait posts in social media have been recently studied~\cite{volkova2017separating}. We built on earlier work and extend models developed to detect four types of deceptive posts on Twitter into regression models capable of predicting how clickbaity the Twitter post is on a scale from 0 to 1.

Earlier work on detecting whether a message is a clickbait or not presented a baseline approach using a set of 215 hand-crafted features and the state-of-the-art machine learning classifiers e.g., Logistic Regression, Naive Base and Random Forest~\cite{potthast:2016}. The best performing model achieved 0.79 ROC-AUC evaluated on 2,000 tweets.

Unlike the baseline approach, we did not rely on hand-crafted features. Instead, we learned representations from images and text and used them to train neural network models. In addition, we used linguistic resources to automatically extract biased language markers to further enrich our models with linguistic cues of uncertainty.  

Clickbait posts are not the only type of deceptive content propagated online and in social media. Recent work focused on automatically detecting other types of deceptive information going beyond clickbait detection e.g., propaganda, satire~\cite{Rubin:15}, hoaxes and disinformation in web-pages~\cite{perez2017automatic,rashkin2017truth,wang2017liar} and social media~\cite{volkova2017separating}.  

\begin{figure*}
\centering
\includegraphics[width=0.47\textwidth]{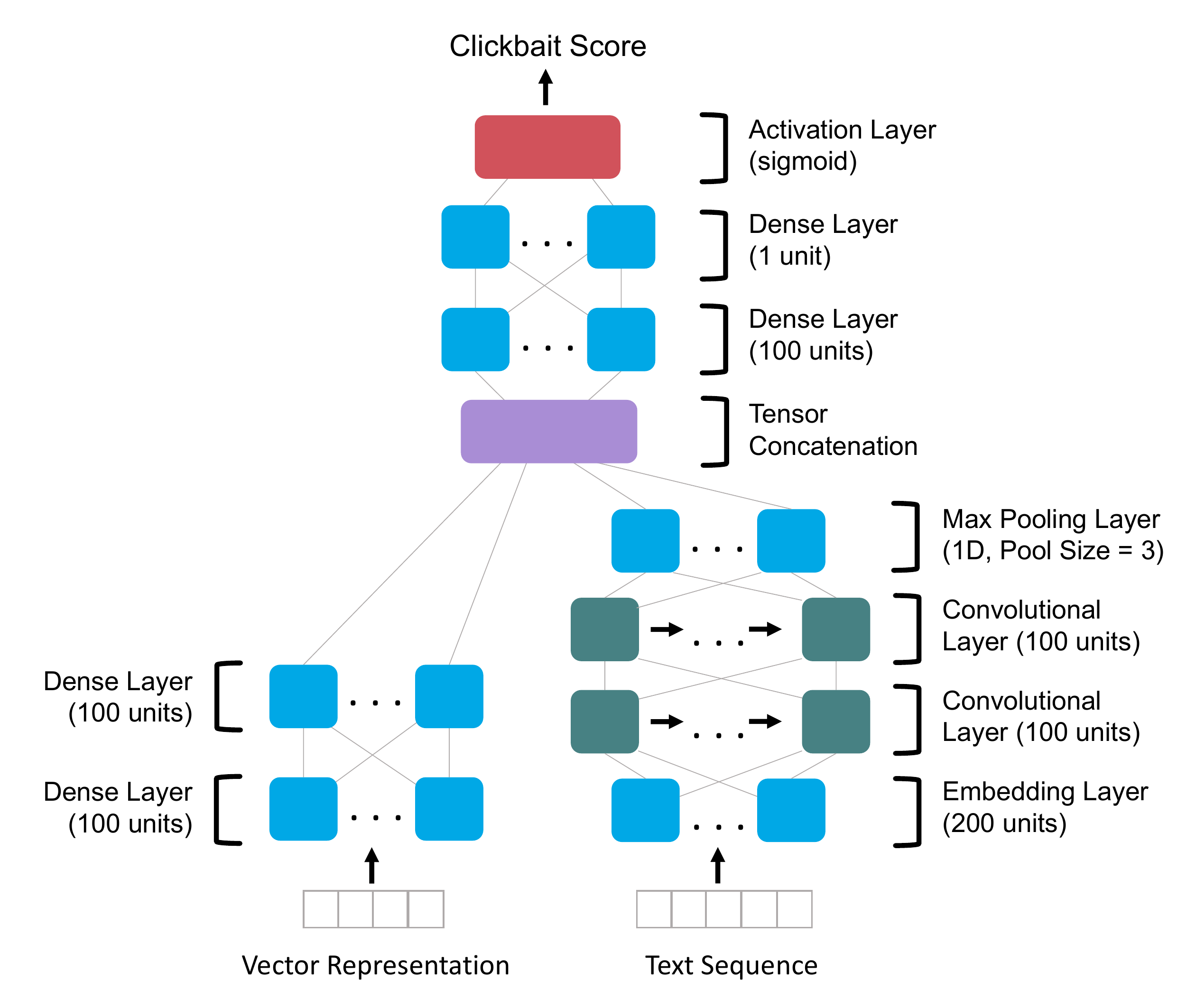} 
\includegraphics[width=0.47\textwidth]{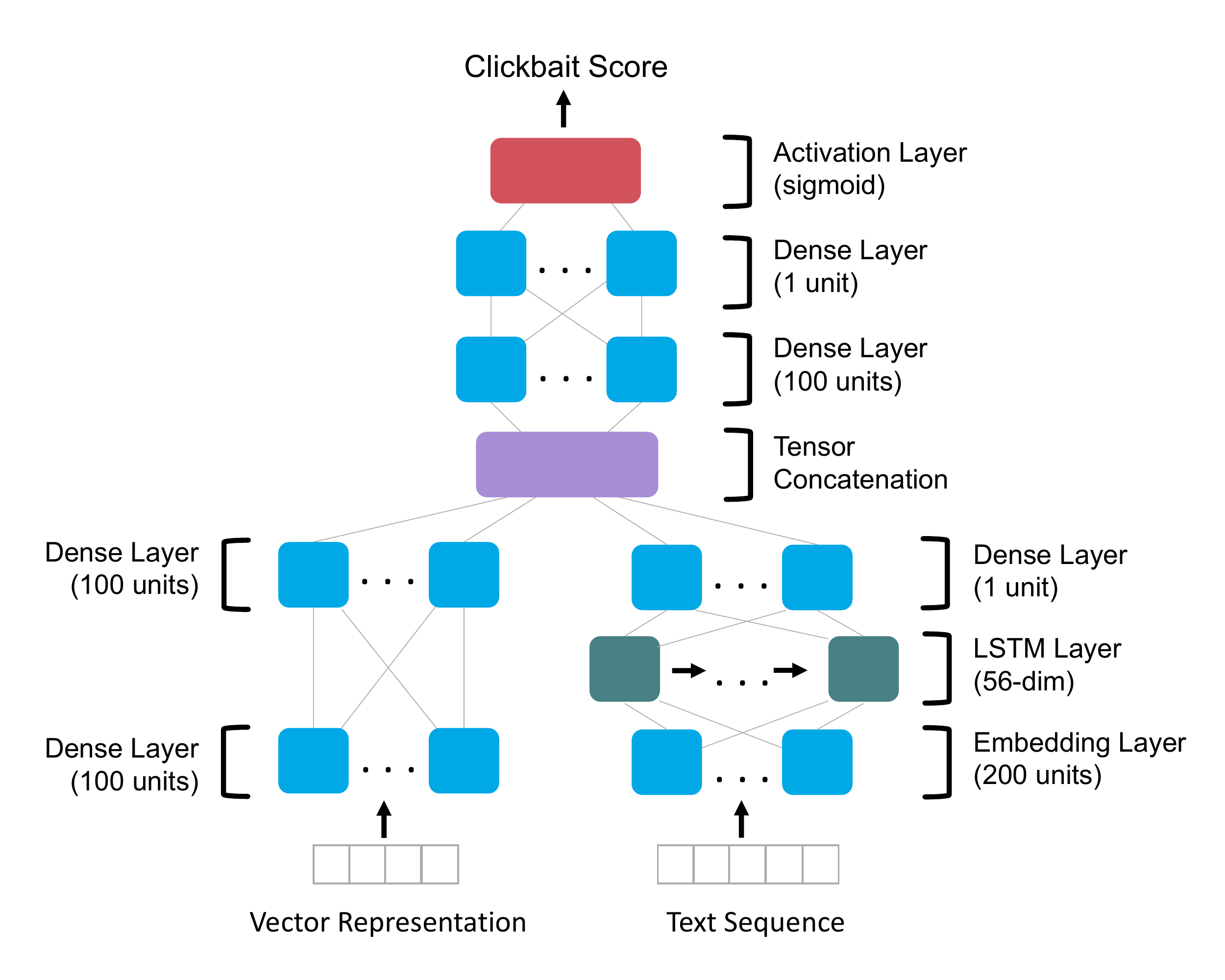}
\caption{CNN (left) and LSTM (right) neural network architectures for clickbait scoring models.} 
\label{fig:model_architecture}
\vspace{-0.3cm}
\end{figure*}

\section{Clickbait Challenge 2017}
For the Clickbait Challenge 2017~\cite{potthast:2017a}, the task is to develop a regressor that provides a score of how clickbait-y a social media post is. In the datasets provided, the content of the post (text, media, and timestamp) and the primary content of the linked web page is included for each post. Scores for these posts were calculated as the mean judgment of at least five annotators who judge the post on a 4-point scale as not click baiting (0.0), slightly click baiting (0.3) considerably click baiting (0.66), or heavily click baiting (1.0). Submissions to the challenge were judged based on the mean squared error (MSE) of their score predictions, relative to the mean judgments of the crowd-sourced annotations~\cite{potthast:2017b}. Thus, the task is a regression problem, optimizing MSE.

\begin{table}[b!]
    \centering
    \begin{tabular}{lrrr}
    Dataset & \# posts & \# Clickbait: \# Not \\
    \hline
    2k Labelled & 2,495 & 1:2.23\\
    20k Labelled & 19,538 & 1:3.10\\
    Unlabelled & 80,012 & ?:?\\
    \end{tabular}
    \caption{Sizes and ratios of clickbait to non-clickbait posts of the datasets provided.}
    \label{tab:datasets}
\end{table}

The clickbait challenge organizers provided three datasets for use to build our models~\cite{potthast:2017b}: two labelled datasets and one unlabelled dataset. We will refer to the two labelled datasets by their approximate size, using "2k labelled" to reference the smaller labelled dataset ('clickbait16-train-170331') and "20k labelled" for the larger labelled dataset ('clickbait17-train-170630). As there is only one unlabelled dataset ('clickbait17-unlabeled-170429'), we simply refer to it as the unlabelled data.
Table~\ref{tab:datasets} summarizes the exact size of each of the datasets and the ratios of clickbait to non-clickbait posts in each. Clickbait posts are those with a score of 0.5 or greater and non-clickbait posts are those with a score below 0.5.

\section{Approach}
We use linguistically-infused neural network models that learn strength of clickbait content from the text present not only in the tweets themselves but also the linked articles, as well as images present in the tweets. We hypothesized that while we should be able to get a certain level of performance using the tweet text or images alone, including the linked article content or a joint representation of the tweet and linked article would lead to a significant boost in performance. 

Intuitively, even using a human to judge the strength of click-baiting in a post, the tweet content is a significant factor in determining how clickbait-y it is. Posts such as {\it "This Tumblr account will nail your personality down in a second"}, {\it "The most diverse place in America? It's not where you think"}, or {\it "Once you consume them, they can move throughout your body -- most commonly your brain"} are clearly clickbait. However, other posts can be harder to determine from the tweet alone but considerably easier when you include the linked article and use the inconsistency between what the title hints at and what the article is actually about. For example, {\it "Preserved dinosaur cells. Could scientists recreate a prehistoric beast?"} which links to the article {\it "Found: Preserved dinosaur cells -- but sadly scientists still can't build 'Jurassic World'"}. In this section, we describe neural network architectures and text, linguistic, and image signals we compare the performance of combinations of in our experimental results section.

\subsection{Models}
Linguistically-infused neural network models have been previously used to effectively classify tweets as containing trusted or suspicious content and  to classify tweets with suspicious content into one of multiple classes of deceptive content, clickbait being one such class~\cite{volkova2017separating}. We adapt these models for the regression task and unexplored inputs that we are provided with e.g., images. Two neural network architectures are shown in Figure~\ref{fig:model_architecture}. We rely on state-of-the-art Long Short-Term Memory (LSTM) and Convolutional Neural Network (CNN) layers effectively used in text classification~\cite{johnson2014effective, zhang2015sensitivity}. The text sequence sub-network consists of an embedding layer and either two 1-dimensional convolution layers followed by a max-pooling layer or a 56-dimensional LSTM layer followed by a dense layer. We incorporate additional vector representations of the posts and associated content through the left sub-network. In this sub-network, we include signals from images in the post and/or linguistic cues of the associated text, \textit{i.e.} the linguistic markers of bias present.

\begin{table*}[t!]
    \centering
    \small
    \begin{tabular}{lllrrrrrr}
\sc Model	& \sc	Text Input	& \sc	Vector Input	& \sc	\# epochs	& \sc	MSE	&	\sc RMSE	&\sc	MAE	\\
\hline
AB	&	Tweet	&	--	&	N/A	&\textbf{	0.0513	}&	0.2266	&	0.1830	\\
AB	&	Tweet	&	LC (Tweet)	&	N/A	&	0.0550	&	0.2346	&	0.1868	\\
AB	&	Tweet + Article	&	--	&	N/A	&	0.0570	&	0.2387	&	0.1909	\\
AB	&	Tweet + Article	&	LC (Tweet + Article)	&	N/A	&	0.0570	&	0.2387	&	0.1910	\\
\\													
CNN	&	Tweet	&	--	&	3	& \textbf{	0.0476	}&	0.2183	&	0.1727	\\
CNN	&	Tweet	&	LC (Tweet + Article)	&	3	&	0.0503	&	0.2242	&	0.1756	\\
CNN	&	Tweet	&	LC (Tweet + Article) + 2,048-dim image vectors	&	3	&	0.0491	&	0.2216	&	0.1727	\\
CNN	&	Tweet	&	--	&	5	&	0.0584	&	0.2416	&	0.1898	\\
CNN	&	Tweet	&	LC (Tweet + Article)	&	5	&	0.0514	&	0.2266	&	0.1783	\\
CNN	&	Tweet	&	LC (Tweet + Article) + 2,048-dim image vectors	&	5	&	0.0504	&	0.2245	&	0.1767	\\
CNN	&	Tweet	&	--	&	10	&	0.0612	&	0.2474	&	0.1940	\\
CNN	&	Tweet	&	LC (Tweet + Article)	&	10	&	0.0532	&	0.2307	&	0.1810	\\
CNN	&	Tweet	&	LC (Tweet + Article) + 2,048-dim image vectors	&	10	&	0.0531	&	0.2305	&	0.1811	\\
CNN	&	Tweet	&	--	&	25	&	0.0642	&	0.2535	&	0.1994	\\
CNN	&	Tweet	&	LC (Tweet + Article)	&	25	&	0.0548	&	0.2340	&	0.1851	\\
CNN	&	Tweet	&	LC (Tweet + Article) + 2,048-dim image vectors	&	25	&	0.0543	&	0.2329	&	0.1837	\\
\\													
LSTM	&	Tweet	&	--	&	3	&	0.0471	&	0.2170	&	0.1702	\\
LSTM	&	Tweet	&	LC (Tweet + Article)	&	3	&	0.0449	&	0.2120	&	0.1670	\\
LSTM	&	Tweet	&	LC (Tweet + Article) + 2,048-dim image vectors	&	3	&\textbf{	0.0444	}&	0.2107	&	0.1664	\\
LSTM	&	Tweet	&	--	&	5	&	0.0532	&	0.2307	&	0.1810	\\
LSTM	&	Tweet	&	LC (Tweet + Article)	&	5	&	0.0525	&	0.2290	&	0.1803	\\
LSTM	&	Tweet	&	LC (Tweet + Article) + 2,048-dim image vectors	&	5	&	0.0515	&	0.2269	&	0.1789	\\
LSTM	&	Tweet	&	--	&	10	&	0.0619	&	0.2488	&	0.1963	\\
LSTM	&	Tweet	&	LC (Tweet + Article)	&	10	&	0.0568	&	0.2384	&	0.1879	\\
LSTM	&	Tweet	&	LC (Tweet + Article) + 2,048-dim image vectors	&	10	&	0.0555	&	0.2355	&	0.1856	\\
LSTM	&	Tweet	&	--	&	25	&	0.0648	&	0.2546	&	0.2001	\\
LSTM	&	Tweet	&	LC (Tweet + Article)	&	25	&	0.0618	&	0.2486	&	0.1963	\\
LSTM	&	Tweet	&	LC (Tweet + Article) + 2,048-dim image vectors	&	25	&	0.0609	&	0.2467	&	0.1946	\\
\hline
    \end{tabular}
    \caption{Results of AdaBoost (AB) baselines and different combinations of models and inputs when trained on the 20k labeled dataset and tested on the 2k labeled dataset provided. Results are separated by model architecture used (AB, CNN, or LSTM) and sorted by the number of epochs used in training. Lowest MSE for each model is highlighted in bold. Text only models (using only the right sub-network of the LSTM or CNN model architectures shown above) are indicated with a $-$ in the Vector Input column.}
    \label{tab:results_combinations_table}
    \vspace{-0.3cm}
\end{table*}

We initialize our embedding layer with pre-trained GloVe embeddings~\cite{glove}, using the 200-dimensional embeddings trained on Twitter. In order to incorporate linguistic cues and/or image vector representations into our networks in addition  text representations, we use the ``late fusion'' approach {that has been effectively used in vision tasks}~\cite{karpathy2014large,park2016combining}. ``Fusion'' allows for a network to learn a combined representation of multiple input streams and can be done early (in the feature extraction layers) or later (\textit{e.g.,} in classification or regression layers). Using fusion as a technique for training networks to learn how to combine data representations from different modalities (text sequences and vector representations) to boost performance at a later stage in our model, we concatenate the output from the content and linguistic or image sub-networks. We hold out the 2k labelled dataset and train each neural network model architecture and input combination on the larger, 20k labelled dataset for a range of epochs (from 3 to 25) using the ADAM optimizer~\cite{kingma-ba-adam}. We use AdaBoost (AB) regressor with unigrams and tf-idf vectors for each input combinations as baselines to compare against.

\subsection{Text Representations}
For each post in the dataset, the text content of the post itself and the associated article, the title, keywords, description, any figure captions, and the actual paragraphs of the article itself, are provided.  We compare the performance of models that use the text of the post only, of the associated article only, and of both the post and article combined into a single document, where the text of the article is defined as the title, keywords, description, and paragraphs. 

For our neural network models, we process the raw text inputs into padded text sequences using the following process. First, we remove hashtags, mentions, punctuation, and urls. Then, we tokenize each input string using the keras\footnote{https://keras.io/} text pre-processing tokenizer with a maximum of 10,000 words that was fit on the entire corpus (training and test data) so that our models would use a shared representation of the training and testing data in the embedding layer. These text sequences are truncated, or padded with zeros, to 100-dimensional vectors. These 100-dimensional vector representations of the text sequences are then passed into the model (right hand side of the LSTM or CNN architectures illustrated in Fig.~\ref{fig:model_architecture}).

\subsection{Linguistic Cues of Uncertainty}
We use the linguistic cues found in the content of the tweet and the title, keywords, description, and paragraphs of the associated article to boost performance of our models. Previous work identified the use of linguistic cues of biased languages to identify clickbait~\cite{volkova2017separating}. Thus, we extract {\it assertive verbs} (which indicate the level of certainty in the complement clause)~\cite{Hooper:75}, {\it factive verbs} (which presuppose the truth of their complement clause)~\cite{Kiparsky:68fact}, {\it hedges} (expressions of tentativeness and possibility)~\cite{Hyland:05metadiscourse}, {\it implicative verbs} (which imply the veracity of their complement)~\cite{Karttunen:71implicative}, and {\it report verbs}~\cite{Recasens:13} from pre-processed text. Similar linguistic cues have also been used to identify biased language on Wikipedia~\cite{Recasens:13}.

\subsection{Image Representations}
We relied on ResNet architecture~\cite{he2016deep} initialized with ImageNet weights~\cite{deng2009imagenet} to extract 2,048-dimensional feature vectors from 10k Twitter clickbait images in the labeled sets. We used the image vector representations as an input to a neural network model as shown in Figure~\ref{fig:model_architecture}.

Additionally, we used Google's Tensorflow Object Detection API\footnote{https://github.com/tensorflow/models/tree/master/research/object\_detection} to extract objects from the clickbait images. Several models have been made available with varying speed and accuracy trade-offs. All models were pre-trained on the Common Objects in Context (COCO)\footnote{http://cocodataset.org/\#home} dataset which contains 80 object classes. Due to the sparsity of COCO objects appearing in our images, we chose to implement the detection model with the highest performance even though it requires a longer run time. The model consists of a Faster R-CNN meta-architecture and utilizes Inception Resnet (v2) as a feature extractor~\cite{object-detection}. Object tags were not included in the final deployment of our model due to time and computation constraints. However, we performed the analysis of objects that appear on clickbait vs. non-clickbait Twitter post images as reported in Section~\ref{sec:image_analysis}.

\section{Experimental Results}  
We experiment with different combinations of inputs to the neural network architectures in Figure~\ref{fig:model_architecture}. We compare the results of each combination of inputs and model architectures, training on the 20k labelled dataset, using 20\% of the data as validation data, and testing on the 2k labelled dataset. Although our task is to minimize the mean squared error (MSE) of the clickbait scores predicted, we also consider the root mean squared error (RMSE) and the mean absolute error (MAE) of our models' predicted scores when comparing model performance.

After comparing the performance of predicting clickbait scores using different types of inputs, we found the highest performance from models that use the tweet as the text sequence input combined with the linguistic cues (LC) extracted from the tweet and the title, keywords, description, and paragraphs of the linked article. We report the results of these input combinations for the LSTM and CNN architectures in Table~\ref{tab:results_combinations_table}, along with the results of our baseline Adaboost (AB) models. For each neural network model we report the results after training for 3, 5, 10, and 25 epochs. Results are ordered by model, inputs and the number of epochs. 

We show that the best performing models were trained on fewer epochs (3 or 5) and that the worst performing models were the models using text only inputs and 25 epochs. Thus, we focused on models training on 3 and 5 epochs. Although the models that included the 2,048-dimensional image vectors performed slightly higher than equivalent models without, e.g. the best LSTM models with images (MSE = 0.0444) and without images (MSE = 0.0449), they do not perform significantly better. As there is a much longer run time for models that need to extract the 2,048-dimensional image vector representations and no significant increase, we focus on the best models without the image representations for this challenge.

\subsection{Incorporating Noisy Labels}
Although we were not able to deploy these models before the end of the challenge submission period, we also developed models trained with noisy labels as well as the clean labels provided in the 20k and 2k labelled dataset. To do this, we took one of our top performing models (MSE=0.0449), the LSTM model that used the text sequence input from the tweets and the linguistic cues vector representations of the tweet and article documents, and labelled the 80k unlabelled tweets provided. Then we combined the 20k labelled dataset and the 80k unlabelled dataset with our noisy labels our model predicted into a single dataset of approximately 100k examples, which we refer to as the noisy label dataset. The ratio of this combined dataset is skewed slightly more to non-clickbait posts than the 20k labelled dataset had, as shown in Table~\ref{tab:noisy_data_splits}.

\begin{table}[htb]
    \centering
    \begin{tabular}{lrrr}
    \sc Dataset &  \sc\# Posts & \sc \# Clickbait: \sc \# Not \\
    \hline 
    Unlabelled & 80,012 & 1:3.98\\
    Noisy Labelled & 99,551 & 1:3.78\\
    \end{tabular}
    \caption{Ratios of clickbait to non-clickbait labelled posts of the noisily labels for the unlabelled dataset and the noisy labelled dataset that incorporates both the 20k labelled dataset and the noisy labels for the unlabelled data.}
    \label{tab:noisy_data_splits}
        \vspace{-0.2cm}
\end{table}

Next, we compare different combinations of models trained on this noisy dataset. As before, we use a validation split of 80-20 during training, so each model is trained on approximately 80 thousand examples and validated on 20 thousand examples before being tested on the 2k labelled dataset. We report the results of the LSTM and CNN models that use the top performing input combinations on this noisy labelled data in Table~\ref{tab:noisy_label_results}. However, we see that these models did not outperform our previous models trained on the 20k labelled data. Because of the computation time needed to extract the 2048-dimensional image vectors for each image referenced in the unlabelled dataset and the lack of significant performance differences we saw in the previous tables, we do not include results of models including image vector representations. 

\begin{table}[ht]
    \centering
    \small
    \begin{tabular}{l@{\hskip4pt}l@{\hskip5pt}l@{\hskip3pt}r@{\hskip5pt}r@{\hskip5pt}r@{\hskip5pt}r@{\hskip10pt}r@{\hskip10pt}r}
\sc Model	& \sc	Text 	&	\sc Vector 	&\sc	\# epochs	&\sc	MSE	& \sc	RMSE	&	\sc MAE	\\
\hline
CNN	&	tweet	&	--	&	3	&	0.0449	&	0.2119	&	0.1677	\\
CNN	&	tweet	&	LC (tweet+article)	&	3	&\textbf{	0.0435	}&	0.2086	&	0.1647	\\
CNN	&	tweet	&	--	&	5	&	0.0460	&	0.2145	&	0.1696	\\
CNN	&	tweet	&	LC (tweet+article)	&	5	&	0.0439	&	0.2095	&	0.1657	\\
CNN	&	tweet	&	--	&	10	&	0.0478	&	0.2186	&	0.1721	\\
CNN	&	tweet	&	LC (tweet+article)	&	10	&	0.0447	&	0.2114	&	0.1674	\\
CNN	&	tweet	&	--	&	25	&	0.0476	&	0.2181	&	0.1726	\\
CNN	&	tweet	&	LC (tweet+article)	&	25	&	0.0442	&	0.2102	&	0.1666	\\ \\
LSTM	&	tweet	&	--	&	3	&	0.0450	&	0.2120	&	0.1665	\\
LSTM	&	tweet	&	LC (tweet+article)	&	3	&\textbf{	0.0448	}&	0.2116	&	0.1664	\\
LSTM	&	tweet	&	--	&	5	&	0.0451	&	0.2123	&	0.1666	\\
LSTM	&	tweet	&	LC (tweet+article)	&	5	&	0.0458	&	0.2140	&	0.1679	\\
LSTM	&	tweet	&	--	&	10	&	0.0459	&	0.2143	&	0.1685	\\
LSTM	&	tweet	&	LC (tweet+article)	&	10	&	0.0454	&	0.2130	&	0.1679	\\
LSTM	&	tweet	&	--	&	25	&	0.0469	&	0.2166	&	0.1705	\\
LSTM	&	tweet	&	LC (tweet+article)	&	25	&	0.0470 &	0.2168	&	0.1705	\\
\hline
    \end{tabular}
    \caption{Results of neural network models trained on the noisy labelled dataset and tested on the 2k labelled dataset. Results are separated by model architecture used (CNN, or LSTM) and sorted by the input combinations and the number of epochs used in training. Lowest MSE for each model is highlighted in bold. Text only models (using only the right sub-network of the LSTM or CNN model architectures shown above) are indicated with a $-$ in the Vector Input column. }
    \label{tab:noisy_label_results}
    \vspace{-0.3cm}
\end{table} 

\begin{figure}
\centering
\includegraphics[width=0.45\textwidth]{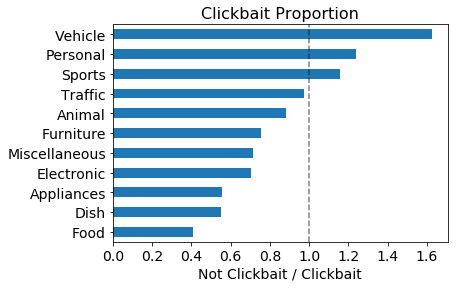} 
\caption{Proportion of objects extracted from non-clickbait to clickbait images clustered into eleven COCO categories.} 
\label{fig:scatter}
\vspace{-0.3cm}
\end{figure}

\subsection{Analyzing Objects  in Clickbait Images}
\label{sec:image_analysis}
\begin{figure}
\centering
\includegraphics[width=0.48\textwidth]{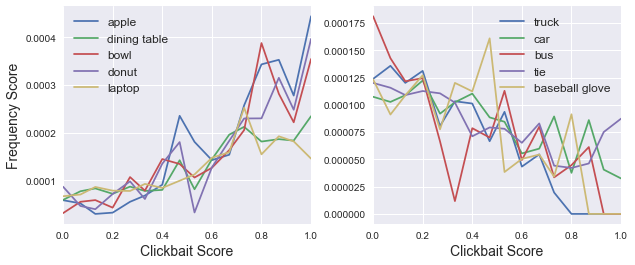} 
\caption{Trends in object proportions as a function of the clickbait score.} 
\label{fig:dist}
\vspace{-0.3cm}
\end{figure}

Figure~\ref{fig:scatter} reports preliminary results on objects extracted using Tensorflow Object Detection API from clickbait vs. non-clickbait images. We found that only 10,270 (47\%) posts out of 22,033 tweets in labeled dataset contained images. We were able to extract objects from 9,488 out of 10,270 images (2,057 clickbaits and 7,431 non-clickbaits). We grouped objects extracted from images into eleven COCO categories. We found that there are more objects from vehicle, sports and personal categories in non-clickbait images than in clickbait images. There are no significant differences in objects from furniture, animal, and traffic categories. We observed that clickbait images have more objects from food, appliances, and dish categories compared to non-clickbait images. Figure~\ref{fig:dist} presents trends for object proportions as a function of the clickbait score. We observe that the number of food objects e.g., apple, donut; furniture objects e.g., dining table and electronics objects e.g., laptop increases as the clickbait score increases; and the number of vehicle objects e.g., car, bus, truck decreases as the clickbait score increases.

\section{Conclusions} 
We proposed a neural network architecture that used the text of the tweet, the linguistic cues found in the tweet and associated article (e.g., the text of the tweet, article title, article keywords, article description, and article paragraphs) to compete in the Clickbait Challenge 2017. Our model for predicting a clickbait score for a given tweet performed as well or better than the other combinations of features and model architectures that we compared. Among the top-performing models, LSTM-based model also had the most reasonable run-time and computational requirements to be deployed in the challenge. We trained this model on the 20k labelled dataset and came in 5th out of the 13 other clickbait detectors\footnote{http://www.tira.io/task/clickbait-detection/dataset/clickbait17-test-170720/} with a mean squared error of 0.04 and an $R^2$ of 0.44. The final test sample contained 18,979 instances with a 1:3.2 clickbait to not-clickbait ratio, similar to the ratio of 1:3.10 of the 20k labelled dataset. This is a similar performance to that which we saw when testing on the 2k labelled dataset where we also saw a MSE of 0.04, as noted in Table~\ref{tab:results_combinations_table}.  

\section{Future Work}
In recent work~\cite{orderEmbeddings}, a visual-semantic hierarchy is used to learn ordered embeddings to model the relationship between images and their descriptions. To capture relationships between posts and their associated images, we would like to apply this approach to the clickbait posts and their associated images in order to extract more precise features from the tweets, capable of demonstrating how revealing an image is of its article's content.

In addition, we would like to incorporate joint representations of the relationships between different components of clickbait posts in the vector representations. For example, we would consider the edit distances or Doc2Vec similarities of the post text and article titles as a way to capture the relationship between post and article. 

Although we did not include  objects extracted from images in our final submission, our initial results indicate that including objects extracted from associated images could boost performance. 
Finally, with the imbalance of classes within the data, model performance could also be improved with more extensive use of techniques to deal with skewed data than we employed in this study.\footnote{http://www.kdnuggets.com/2017/06/7-techniques-handle-imbalanced-data.html}

\begin{raggedright}
\bibliography{clickbait17} 
\end{raggedright}
\end{document}